\documentclass[sigconf]{acmart}
\renewcommand\footnotetextcopyrightpermission[1]{} 
\usepackage{booktabs} 
\usepackage{graphicx}
\usepackage{hyperref}
\usepackage{gensymb}
\usepackage{amsmath}
\usepackage{amssymb}
\usepackage{enumerate}
\newtheorem{definition}{Definition}

\newcommand{\gmle}{\hat{\gamma}_{MLE}}
\newcommand{\gmlei}{\hat{\gamma}_{MLE,i}}
\newcommand{\gmlej}{\hat{\gamma}_{MLE,j}}
\newcommand{\gjs}{\hat{\gamma}_{JS}}
\newcommand{\gshr}{\hat{\gamma}_{SHR}}

\newcommand{\1}{\mathbf{1}}

\newcommand{\ep}{\mathbf{\epsilon}}
\newcommand{\diag}{\text{diag}}
\newcommand{\J}{\mathcal{J}}
\newcommand{\E}{\mathbb{E}}
\newcommand{\I}{\mathcal{I}}
\newcommand{\D}{\mathcal{D}}

\newcommand{\nba}{{\tt NBA2016} }
\newcommand{\nfl}{{\tt NFL2016} }

\acmConference[]{Arxiv}{July 2018}{}
\acmYear{2018}
\copyrightyear{2018}

\begin{document}

\author{Stephen Ragain}
\affiliation{%
  \institution{Stanford University}
  \streetaddress{475 Via Ortega}
  \city{Stanford} 
  \state{CA} 
  \postcode{94305}
}
\email{sragain@stanford.edu}

\author{Alexander Peysakhovich}
\affiliation{%
  \institution{Facebook Artificial Intelligence Research}
}
\email{alex.peys@fb.com}

\author{Johan Ugander}
\affiliation{%
  \institution{Stanford University}
  \streetaddress{475 Via Ortega}
  \city{Stanford} 
  \state{CA} 
  \postcode{94305}
}
\email{jugander@stanford.edu}

\title{Improving pairwise comparison models using Empirical Bayes shrinkage}

\begin{abstract}
Comparison data arises in many important contexts, e.g.~shopping, web clicks, or sports competitions. Typically we are given a dataset of comparisons and wish to train a model to make predictions about the outcome of unseen comparisons. 
In many cases available datasets have relatively few comparisons (e.g.~there are only so many NFL games per year) or efficiency is important (e.g.~we want to quickly estimate the relative appeal of a product). In such settings it is well known that shrinkage estimators outperform maximum likelihood estimators. A complicating matter is that standard comparison models such as the conditional multinomial logit model are only models of conditional outcomes (who wins) and not of comparisons themselves (who competes). As such, different models of the comparison process lead to different shrinkage estimators.
In this work we derive a collection of methods for estimating the pairwise uncertainty of pairwise predictions based on different assumptions about the comparison process. These uncertainty estimates allow us both to examine model uncertainty as well as perform Empirical Bayes shrinkage estimation of the model parameters. We demonstrate that our shrunk estimators outperform standard maximum likelihood methods on real comparison data from online comparison surveys as well as from several sports contexts.
\end{abstract}

\maketitle

\section{Introduction}
Comparison-based choice data, where one item is selected from a choice set, is ubiquitous in both online and offline applications. The use of past choices to predict future choices is at the heart of online commerce (e.g.~recommender systems \cite{schafer1999recommender}),  forecasting competitions (e.g.~Elo ratings in games like chess \cite{elo1978rating}), and social science (e.g.~modeling demand systems in economics \cite{carlton2015modern}).  A popular workhorse for this task is the conditional multinomial logit (MNL) model \cite{luce1959individual,luce1977choice}, which is usually estimated using maximum likelihood methods \cite{hunter2004mm,krishnapuram2005sparse}. 

It is well known that maximum likelihood methods are typically consistent (that is, learn the true parameters as the amount of data goes to infinity), but can have high variance. This can be a problem in datasets where the effective number of data points per model parameter is small. For example, we may wish to learn a new chess competitor's rating quickly, or our dataset may be restricted to be of a finite size (each consumer only makes a fixed amount of choices per week), or certain comparisons may be more likely than others (such as in sports leagues with conference structures). 

In such ``small data'' situations, shrunken (regularized) estimators often perform much better than maximum likelihood estimators. In a classic example, Efron and Morris \cite{efron1973stein} showed that if we want to estimate a baseball batter's skill from a sample of at-bats, we should shrink each batter's estimated skill independently towards the global average by an amount that is proportional to the variance in our estimate. Procedures like this one are often called {\it Empirical Bayes} \cite{carlin2000bayes}, and this shrinkage can substantially change estimates and also greatly improve out-of-sample prediction.

To apply a shrinkage estimator in practice one needs to know how much to shrink and what to shrink toward. When estimating multiple means simultaneously as Efron and Morris did, the choices are straight-forward. In comparison models, however, the direction and magnitude of shrinkage is intimately tied to model uncertainty, specifically uncertainty about pairwise comparisons and not absolute values of individual parameters. In this paper we propose a family of simple methods for estimating different notions of parameter uncertainty associated with any conditional multinomial logit model, which allows us both to examine the model's weaknesses and to improve parameter estimates via shrinkage.

The MNL model, like many choice models, assumes that observed data results from a stochastic process where items have an underlying ``score.'' The MNL model further assumes that the probability that one item is chosen over another is proportional to these scores (Luce called this the ratio scale representation \cite{luce1959individual}). Given a comparison dataset, the task of the analyst is to estimate these latent scores from observed comparisons and then use them to make subsequent predictions in future comparisons.

Our contribution is to develop an Empirical Bayes-like procedure for shrinkage estimation with the MNL model. In the datasets we will consider each data point as a pair of items and a choice from that pair. This means that the uncertainty in individual parameters (e.g.~scores for each object) is generally intertwined. Furthermore, this uncertainty can be greatly affected by the sampling process. Consider a season of American football (NFL) games, where games can be interpreted as comparisons between teams. In the NFL teams are split into conferences and divisions and play mostly other teams in the same conference. This split means that given a season of data we can be relatively sure how to rank two teams within one conference because we have seen many games within that conference. However, we may at the same time be quite uncertain about the outcome when the two teams are in different conferences because we have relatively few comparisons of teams across conferences. 

We develop a family of procedures for shrinkage estimation with the MNL model. The key intuition behind all our approaches is that unlike in standard statistical models where we focus on the uncertainty of a parameter estimate, we need to focus on pairwise uncertainty among parameters. Towards this goal, we focus on the covariance matrix (or the related Fisher information matrix) of the estimated parameters, but can only do so under some assumptions about the distribution of how pairs are compared. How items are compared is generally not the domain of discrete choice modeling, which models the probability of different outcome of a comparison conditional on the comparison being made, rather than the probability that the comparison is made. We discuss the importance of modeling this distribution in the shrinkage estimators we develop. 

Given a set of assumptions about uncertainty, we adapt the classic James--Stein shrinkage estimator \cite{james1961estimation} to the choice problem as follows: first, we compute the maximum likelihood estimate of the model parameters and a covariance matrix of its parameters, based on a specific notion of uncertainty. We use this (generally non-diagonal) matrix to shrink our maximum likelihood estimates. 
We present four methods of computing a covariance matrix using bootstrap-based methods as well as two using Fisher information. We discuss which methods of covariance are most suited to different methods of data generation and apply our procedures to several real datasets. We find that the properly shrunk estimators have better predictive power than the standard MLE. 

While we focus on the MNL model, being the most popular discrete choice model, similar procedures can be applied to choice models other than MNL such as Thurstonian models \cite{thurstone1927law}, random utility models (RUMs) \cite{soufiani2013preference}, mixed logit \cite{train2009discrete}, nested logit \cite{hausman1984specification}, elimination by aspects \cite{tversky1972elimination}, Markov chain-based models \cite{ragain2016pairwise,blanchet2016markov}, 
the Blade-Chest model \cite{chen2016modeling}, and Elo ratings \cite{elo1978rating}. We leave these extensions to future work.

\section{Choice Models}
Our main object of interest will be a universe $U$ of items, where we denote the items by indexes $\lbrace 1, \dots, n \rbrace$. At training time we will be given a dataset $\mathcal{D}$ consisting of ordered tuples $(i,j)$ representing that $i$ has been chosen over $j$, or equivalently $i$ has been chosen from the set $\{i,j\}$. We focus on shrinking choice models learned from binary choice sets (pairs), but the ideas are easily extensible to the general case of choices from sets of arbitrary and mixed size. At test time we will receive a new dataset of binary choice sets and our goal will be to predict which item will be chosen from each set. We will focus on random utility models.

\begin{definition}
A random utility model of choice consists of two components:
\begin{itemize}
\item A utility vector $\gamma$ of length $n$ where $\gamma_i > 0$ is a underlying numeric ``quality'' of item $i$.
\item A choice rule, which we denote $p_{ij} = f(\gamma_i, \gamma_j)$, that determines the probability $i$ is chosen from $\lbrace i, j \rbrace.$
\end{itemize}
\end{definition}

We will give specific emphasis to two such models in this work: the Bradley-Terry-Luce model (BTL) \cite{bradley1952rank} for pairwise comparisons and the Rasch model \cite{rasch1960probabilistic} of item-response outcomes, both being special cases of the conditional Multnomial Logit (MNL) \cite{mcfadden1973conditional} model (sometimes also called the Plackett-Luce model \cite{plackett1975analysis}).

\begin{definition}
The Bradley-Terry-Luce (BTL) model uses the choice function $$p_{ij} = f(\gamma_i, \gamma_j) = \dfrac{\gamma_i}{\gamma_i + \gamma_j}.$$
\end{definition}

The BTL model is scale-invariant (multiplying $\gamma$ by a scalar factor results in the same probabilities). A common convention is to normalize the parameter vector $\gamma$ such that $|| \gamma ||_1=1$. 

The Rasch model is a special case of the BTL model specifically targeted at comparisons between items from disjoint contexts such as (student, question) pairs in test-taking or (offense, defense) pairs in sports outcomes. The Rasch model divides the universe of alternatives $U$ into two disjoint subsets $U_1,U_2$ with comparisons only between items of different types. As with BTL, the Rasch model is scale invariant. To make it identified one can split the $\gamma$ vector into $\gamma^{U_1}$ and $\gamma^{U_2}$, and then require that $||\gamma^{U_1}||_1 + ||\gamma^{U_2}||_1 = 1$. Because the BTL model is a special case of the MNL and our work generalizes to the MNL model, we will refer to $\gamma$ as MNL parameters hereafter. 

The MNL model can be though of as a choice process where items have real underlying qualities but at choice time Gumbel noise is added to both items' qualities, with choices made based on which item's realized quality is higher \cite{luce1959individual,manski1977structure,yellott1977relationship}. The Gumbel noise can be changed to Gaussian noise to transform the BTL and Rasch models into Thurstone \cite{thurstone1927law} and Lawley-Lord models \cite{lawley1943xxiii}, respectively. 
There is very little practical difference between the BTL/Rasch (logit) and Thurstone/Lawley-Lord (probit) models of discrete choice.

\section{Bias-Variance Tradeoffs in MNL Estimation}
We now discuss the standard method for estimating the parameter vector $\gamma$, maximum likelihood estimation (MLE). Given the choice rule in the MNL model we can write the log-likelihood of $\gamma$ given choice dataset $\mathcal{D}$ with $N$ datapoints and generic entry $i_k$ chosen from $\lbrace i_k, j_k \rbrace$ as:
$$
\ell(\gamma;\mathcal{D}) = \sum_{(i_k,j_k) \in \mathcal{D}} \log(\gamma_{i_k}) - \log(\gamma_{i_k} + \gamma_{j_k}).
$$

It is well known that the MLE exists only under certain conditions on $\mathcal{D}$. When $\mathcal{D}$ contains only pairwise comparisons we can construct the auxiliary variables $M^{\mathcal{D}}_{ij}$ as the number of times $i$ is chosen from $\lbrace i, j \rbrace$, and can consider these variables as a directed graph defined on the set of elements. We call this graph the \textit{comparison graph} of $\mathcal{D}$. There exists a unique maximizer to the MLE problem if and only if the comparison graph of $\mathcal{D}$ is strongly connected \cite{ford1957solution}. There are many algorithms for computing this maximum \cite{dykstra1956note, maystre2015fast}.

When the distinction is relevant we will let $\gmle$ denote the maximum likelihood {\it estimator} while letting $\gmle(\mathcal{D})$ denote the maximum likelihood {\it estimate} for a given dataset $\mathcal{D}$. The MLE is an unbiased estimator, that is, if the model is the true data generating process then $\mathbb{E}[ \gmle(\mathcal{D}) ] = \gamma$. We are however still faced with a bias-variance tradeoff.

A common way to make this bias-variance tradeoff comes from Empirical Bayes estimation. The simplest version is the James--Stein estimator (JS) which takes the MLE estimates and shrinks them toward a fixed vector. Though the JS is a biased estimator, it has lower variance than the MLE and gives better out-of-sample prediction in terms of mean-squared error. For multivariate Gaussian data, the JS estimator has a closed form solution \cite{berger1980robust} given by:

$$
\gjs = (I-\Sigma(\Sigma+A)^{-1})\gmle + \Sigma(\Sigma+A)^{-1}u,
$$ 
where $\Sigma$ is the covariance matrix of $\gmle$ under the data $\mathcal{D}$ and $u$ and $A$ are the respective mean and covariance of the true parameters $\gamma^*$. When the true parameters $\Sigma$ are unknown (which is most of the time) sample estimates of the parameters are used as plug-ins to the equation.

Applying the JS estimator to our problem is made difficult by two complications. First, the process we are modeling is non-Guassian,  as the quality parameters are constrained to live on a simplex, but this is a minor point in practice since we can proceed under the assumption that we are simply modeling the first two moments of the data. More importantly, the covariance matrix $\Sigma$ is tightly connected to a data-generating process that, for our setting, is unknown. Given a choice dataset $\mathcal{D}$ and the corresponding maximum likelihood estimator $\gmle$, we should expect $\gmle$ to have low variance when predicting matchups that occur often in $\D$, and high variance for pairs of items which do not appear often. Choice models such as MNL do not model which comparisons are being model, only the outcome; indeed, this is why MNL is called the {\it conditional} multinomial logit: it predicts outcomes {\it conditional} on what comparisons are made.

Though the estimator learns one parameter per item, the data-generating process can create high amounts of \textit{covariance} between parameters -- for example, if items can be split into two categories (e.g.~two football conferences) then the MLE may deliver a low variance estimate of the probability of a team $x$ beating another team $y$ when both teams come from the same conference, but a high variance estimate of the same probability when teams are from different conferences.

Thus a main question will be how to estimate $\Sigma, A$ and $u$ for comparison data. We will refer to the estimator of $\gamma$ based on estimates of $\Sigma, A$ and $u$ as $\hat{\gamma}_{SHR}$, rather than $\hat{\gamma}_{JS}$, and reserve $\hat{\gamma}_{JS}$ for cases where those quantities are known. 

\subsection{What determines MLE uncertainty?}
\label{sec:bd}
A central thesis of this work is that the MNL parameters $\gamma$ alone do not provide a generative model for data, because they do not give us information about which alternatives will be compared. The data arises both from randomness in the choice rule as well as randomness in what alternatives are compared. 

Consider the symmetric matrix $B^{\D}$ defined as counting the number of times each pair is compared: $B^{\D}_{ij}$ is the number of times $(i,j)$ or $(j,i)$ are in $\D$. In this work we assume that the winners of matchups in $\D$ are independent of which other choice sets appear in $\D$. This means that the probability of observed the data $\D$ can be written as:
$$
Pr(\D) = Pr(B^{\D})\left(\prod_{(i,j) \in \D} p_{ij}\right).
$$
The parameters of the choice model only affect the latter product of choice probabilities, but the variance of the MLE choice parameters for a given $\D$ is also affected by probability of $B^\D$, the matchup structure observed in $\D$. 

In this work, we only consider two very simple models for $Pr(B^\D)$, one of which fixes $B^\D$, fixing which matchups are seen based on the observe data, and the other of which samples matchups with replacement from $\D$. We introduce this notation to highlight that a formalization of the distribution of $\gmle$ requires us to make such an assumption. Whether we can e.g.~leverage domain knowledge to model $Pr(B^\D)$ in some principled way to further improve out-of-sample prediction is an interesting direction for future work, but the best approach is likely domain-dependent. 

\section{Estimating $\Sigma$, $A$, and $u$}
The covariance matrix $\Sigma$ can be estimated using either analytic methods based on asymptotic theory or using bootstrap-based methods. These methods each have different costs and benefits. We will introduce the methods in this section and compare them in experiments with real data. 

\subsection{Fisher Information-based methods}
The Fisher Information of the data $\mathcal{D}$ can be used to estimate 
$\Sigma$. There are two ways to express the Fisher information:

\begin{definition}
The \textbf{observed Fisher information} is a function of the dataset and $\gmle$ and we refer to it as $\mathcal{J} (\gmle, \mathcal{D}).$ It is computed by taking the sample mean of the Hessian of the log-likelihood function evaluated at $\gmle$ with the data $\mathcal{D}$.
\end{definition}

\begin{definition}
The \textbf{expected Fisher information}  is a function of the dataset and $\gmle$ and we refer to it as $\mathcal{I} (\gmle, \mathcal{D}).$ It is calculated by taking the dataset $\mathcal{D}$, keeping the distribution of choice sets but replacing the choices with draws from the MNL model implied by $\gmle.$
\end{definition}

We can think of observed Fisher Information as ``non-parametric,'' in the sense that it depends on the data and the likelihood alone, whereas the expected Fisher information is computed assuming that $\gmle$ is the true model behind the choices in the data. Typically the expected Fisher information for a parametric model is a function of the model parameters alone, e.g.~only $\gamma$, and not the observed data. However, we again note that the choice model parameters $\gamma$ do not control the distribution of which choice sets are observed, so $\mathcal{D}$ implicitly serves as a parameter of the information matrices because we use the distribution of choice sets in $\mathcal{D}$ to estimate the true distribution of choice sets. Here we give the derivations when using the empirical distribution of choices given by $\D$, with a discussion of more general choice distributions given later.

To get an estimate of the covariance matrix $\Sigma$ we simply invert $\J(\gmle, \mathcal{D})$ or $\I(\gmle,\mathcal{D})$ and divide by the number of observed matchups $N = |\D|$. We refer to these estimators as $\hat{\Sigma}_\J$ and $\hat{\Sigma}_\I$, respectively. While some work exists on the asymptotic properties of these estimators \cite{abt1998fisher}, we are mostly interested in the finite-sample case, which has its own issues. In particular, there are issues of numerical stability with the matrix inverse and the potential singularity of the matrix. We discuss ways circumvent inverting the information in the following section. Note that the sparsity pattern of the non-zero entries off the diagonal of both $\I$ and $\J$ are the same as the sparsity pattern of the comparison graph for $\mathcal{D}$, giving us a concrete connection between shrinkage using these matrices and the observed data.

{\bf Implicit shrinkage estimation using Fisher Information.}
For larger $n$, especially for comparisons arising from the Rasch model, the Fisher information may not be invertible. However we do not need to invert an estimator $S$ for $\Sigma^{-1}$ to estimate the shrinkage when $\Sigma$ is invertible. We only need to estimate the matrix
\[
R = \Sigma(\Sigma+A)^{-1}
\]
as $\gamma_{js} = (I-R)\gmle + Ru$ where $u$ and $A$ are the respective mean and covariance of our prior on $\gamma$. We have that 
\[
R^{-1} = (\Sigma+A)\Sigma^{-1} = I+A\Sigma^{-1}
\]
so if we use $S$ is an estimator of $\Sigma^{-1}$ rather than using $S^{-1}$ as an estimator of $\Sigma$, we have that 
\[
\hat R = (\hat R^{-1})^{-1} = (I+AS)^{-1}.
\]
Examples of estimators $S$ for the inverse of the covariance include the number of observed matchups $N$ times the observed Fisher information $\J(\gmle,\D)$ or expected Fisher information $\I(\gmle,\D)$. 

\subsection{Bootstrap-based methods}
Another way to estimate $\Sigma$ is to employ a bootstrapping method to generate $K$ replicates of $\mathcal{D}$, obtain estimates $\gmle^{(1)},\dots, \gmle^{(K)}$, and plug these sample estimates into standard estimators for covariance matrices. The simplest way is to construct each replicate is by sampling items from $\mathcal{D}$ with replacement. This procedure may seem attractive, but will often fail because even though $\mathcal{D}$ may induce a strongly connected graph of comparisons -- a requirement for maximum likelihood estimation -- a replicate of $\mathcal{D}$ may not. Consider our running example of two football conferences: if each conference is strongly connected but there are only two games between conferences, one won by a team from the first conference and one by a team from the other, then the graph of $\D$ is strongly connected but if either of those two cross-conference games is not sampled in a particular replicate, the MLE for that replicate will be undefined. 

When bootstrapping often leads to a comparison graph that is not strongly connected, we can get around the above issue by using a block-bootstrap \cite{chernick2011bootstrap}. In the block-bootstrap procedure we take every choice set $\lbrace i, j \rbrace$ that is represented in $\D$ and construct a replicate dataset by resampling with replacement, for each pair, among the items in $\mathcal{D}$ that compare that pair. 

Note that this procedure has the weakness that if $\mathcal{D}$ contains only a single instance of a choice from $\lbrace i, j \rbrace$, or if all of the choices between the pair are the same, those same choices appear in all blocked non-parametric bootstraps. Because we expect Empirical Bayes methods to be applied specifically in domains with data constraints, we will consider a parametric bootstrap as a way around these issues.

Given a model $\gmle$ we construct a parametric bootstrap replicate by taking the dataset $\D$ and replacing the actual choices observed for each entry with a sample from the MNL model with parameters $\gmle.$ Here we can choose to either fix the pairs that are compared -- fixing $B^\D$ from Section~\ref{sec:bd} -- or sample with replacement from the pairs compared in $\D$. 
 
Thus deciding whether or not to block the data on the pairs and whether or not to use parametric methods as part of the resampling scheme gives us four possible approaches to bootstrapping choice data: blocked and parametric (b,p), blocked and non-parametric (b,np), non-blocked and parametric (nb,p), and non-blocked and non-parametric (nb,np).

{\bf Shrinking the variance estimate.}
While $\gmle$ is a consistent estimator in finite samples, the sample covariance matrix generated by the bootstrapping procedures above can be sensitive to outliers. The canonical estimator to first consider for $\Sigma$ is the sample covariance matrix $\hat \Sigma_S$ is 
\[
\hat \Sigma_S = \frac{1}{K-1} \sum_{k=1}^K(\gmle^{(k)} - \bar{\hat{\gamma}}_{MLE})^T(\gmle^{(k)} - \bar{\hat{\gamma}}_{MLE}) 
\]
where $\bar{\hat{\gamma}}_{MLE}$ is simply the mean $K^{-1}\sum_k \gmle^{(k)}$ of the sample MLE vectors. 

 In these cases we can produce more effective estimators of $\Sigma$ by shrinking the sample covariance as well, toward a diagonal matrix of the mean variance. We will use the estimator introduced by \cite{ledoit2004well}:
$$
\hat \Sigma_{SHR} = (1-\nu) \hat \Sigma_S + \nu \bar{\hat{\sigma}}I
$$
where $\bar{\hat{\sigma}}$ is the (scalar) mean sample variance $\bar{\hat{\sigma}} = \frac{1}{n}\sum_{i=1}^n \hat \sigma_{i}$. The shrinkage factor $\nu$ can be tuned by cross validation or one can methods such as those proposed by Ledoit and Wolf \cite{ledoit2004well}. The Ledoit-Wolf method chooses the shrinkage factor to minimize the mean squared error in the covariance matrix, but note that we are here not interested in accurately estimating the covariance matrix other than in the service of then performing shrinkage on the quality parameters $\gamma$. The problem of shrinking estimates of both the location and covariance simultaneously  with the goal of minimize the mean squared error in the location parameters is a difficult problem known as {\it double shrinkage} \cite{zhao2010double}. Optimal double shrinkage estimators are only known for problems with diagonal covariance matrices, whereas non-diagonal covariance is fundamental to our approach. As a result, we employ and recommend selecting this shrinkage factor using cross-validation. 

We find that using this shrunk variance estimate $\hat{\Sigma}_{SHR}$ improves performance over the sample covariance in every application we consider. As a result, in our empirical results we always apply this Ledoit-Wolf shrinkage when estimating $\Sigma$ through bootstrapping. 

\subsection{Which bootstrap? Which information matrix?}

The bootstrap procedures are simpler to understand and the best bootstrap methods outperform Fisher information based methods in our experiments, but bootstrapping is more computationally costly and in certain datasets finding bootstraps which are strongly connected can be intractable, and rejecting such samples may inject bias into the estimation of $\Sigma$. When bootstrapping methods are onerous for these reasons, we recommend using Fisher information based methods for shrinkage. 

Another important practical question is whether the bootstraps should resample matchups  according to the parametric distribution given by the MLE rather than the empirical distribution of that matchup. We find that this distinction is especially important in practice. Consider the \nfl dataset, where each team plays 10 of its 16 games against opponents that it only faces once, so the blocked non-parametric bootstrap will fix this game. Further, the other 6 games come from playing each of 3 divisional opponents twice, and can only change when those games are split. As a result, the non-parametric bootstrap is unable to capture the variance in nearly all of the played games, and as a result, gives a poor estimation of $\hat \Sigma$. 

Choosing the correct bootstrap for a given dataset requires careful consideration of the problem at hand. One cue that the blocked bootstrap may be appropriate is that for some domains the sampling structure of pairs is static or roughly static. For example, we find the block bootstrap is appropriate for the dataset of \nfl matchups we study in which the regular season schedule features similar structure year after year. Meanwhile for the {\tt MLB} dataset we study comparing baseball batters and pitchers, there is less regularity as the rotation of starting pitching and batting orders are decided independently from the underlying team schedules and feature significant changes as the season unfolds based on injuries, performances, trades, etc. For that setting we find a bootstrap (non-blocked) is more appropriate.

{\bf Using priors to ensure strongly connected data.}
 A common solution to MNL inference when the comparison graph is not strongly connected such is to employ a Gamma$(\epsilon,1)$ prior on $\gmle$ (ignoring normalization, which does not change the model) \cite{maystre2016choicerank,guiver2009bayesian}, which leads to a Dirichlet$(\epsilon,\epsilon,\dots,\epsilon)$ prior for the normalization of $\gamma$ \cite{caron2012efficient}.

Noting that the conjugate of the aforementioned Dirichlet prior is the multinomial distribution and that the MNL model extends to choice sets of arbitrary size, we can smooth our data to give a well defined $\gamma$ by adding $\epsilon$ ``choices" of each $x \in U$ from the full set of alternatives. Although we have focused on pairwise comparisons, these larger set comparisons can also be modeled by MNL and the inference algorithms we employ uses a Dirichlet prior with $\epsilon= 10^{-6}$. Further details appear in Appendix \ref{sec:appendix}.

\subsection{Estimation of $A$ and $u$}
In order to estimate the covariance of the true $\gamma^*$ in our estimator $\gmle$  we use a Dirichlet prior for $\gamma^*$ centered at $n\cdot \gmle$ where we multiply with $n$ so that the mean is 1, giving a typically unimodal prior which becomes uniform for $\gmle$ near the uniform vector $u$ where $u_i = 1/n$. The resulting covariance $A$ has
\[
A_{ii} = \frac{\gmlei(1-\gmlei)}{n(n+1)}, 
\hspace{0.3cm}
A_{ij} = \frac{\gmlei\gmlej}{n+1}.
\]

Following this procedure $A$ is not strictly diagonal, which is related to the constraint $||\gamma||_1=1$ coupling entries in $\gamma$. As a result this estimator does not shrink estimates of entries $\gamma_i$ independently, especially if $n$ is small. In general we choose to set $u_i = 1/n$ for all $i \in U$, though given some domain knowledge we may choose a different vector towards which to contract $\gmle$.

\section{Experiments}
We now demonstrate in several datasets that analysts can benefit greatly from capturing the pairwise uncertainty of model parameters in choice models. We consider competition datasets from a variety of sports (baseball, basketball, and American football), as well as a large-scale survey of civic priorities taken as comparisons through the wikisurvey platform AllOurIdeas \cite{salganik2015wiki}. We begin by evaluating inference of MNL models from MNL data, where improvements from the James--Stein estimator are guaranteed. We then focus on out-of-sample prediction on real data, predicting the percent of time an alternative is chosen (e.g.~win percentage of a sports team) in the dataset. 

\subsection{Semi-synthetic data}
We wish to measure the improvement in parameter estimation given by shrinkage. However, ground truth parameters are never known for any real-world dataset. Thus, we will begin with a semi-synthetic data. To construct our dataset, we begin with a real matchup structure from our \nfl dataset (discussed further in the next section) that contains all of the games in the 2016 NFL regular season. This season consists of $N=256$ games played between $n=32$ teams (16 games per team). There is a multi-year rotation of NFL schedules, all of which provide poor connectivity between the two conferences (the NFC and the AFC) \cite{nflcrypt}. 

Next, we randomly generate ``skill'' parameters for each team from the uniform distribution on the simplex and then construct win/loss records by sampling from the implied model using the real game schedule. We then fit our models using this synthetic data and see how well we recover parameters as well as predict unseen matchups. 

We consider two metrics. The first is the relative improvement of the MSE with respect to $\gamma^*$ when using the inferred parameters $\gshr$ compared to $\gmle$: 
$$
\alpha = \E\left[\frac{||\gamma^*-\gmle||^2_2-||\gamma^*-\gshr||^2_2}{||\gamma^*-\gmle||^2_2}\right].
$$
We are also concerned with direct improvement of estimation of the pairwise probabilities themselves. For $\gamma,\gamma'$ let 
$$
 ||\gamma-\gamma'||_P = \frac{1}{n^2}\sum_{i,j} \left|\frac{\gamma_i}{\gamma_i+\gamma_j} - \frac{\gamma'_i}{\gamma'_i+\gamma'_j}\right|
 $$
 denote the mean difference in pairwise probabilities between an MNL model with parameters $\gamma$ and with parameters $\gamma'$. Then let
 $$
\beta =  \E\left[\frac{||\gamma^*-\gmle||_P-||\gamma^*-\gshr||_P}{||\gamma^*-\gmle||_P}\right]. 
 $$
When the model is specified correctly, shrinkage can give large increases in accuracy measured both in terms of parameters and in terms of pairwise probabilities. 

Across 1000 random $\gamma^*$ and resamples of the season, we see the average MSE in recovering $\gamma^*$ reduced by an average of 51\% ($\alpha = 0.51$) with a shrunk estimator. using the expected Fisher information $\hat \Sigma_\I$. Likewise we observe a relative average MSE improvement of $12\%$ ($\beta=0.12$) on the pairwise probabilities.

\subsection{Out-of-sample NFL/NBA predictions}

Having confirmed that shrinkage gives better parameter estimates than MLE when the underlying behavior is generated by an MNL model, we now turn to real world data. We use the real win/loss outcomes from the 2016 NFL season. 

To highlight the differences in improvements between datasets of different sizes, we also introduce the \nba dataset, which contains all of the games played in the 2016 NBA season. \nba consists of $N=1260$ games played among $n=30$ teams (82 games per team). Although each team plays the majority of its games within its conference, each team plays two games against each teams in the other conference, providing less sparsity (more connectivity) relative to the \nfl dataset.

\begin{figure}[t]
\small
\begin{center}
\begin{tabular}{ c | c | c | c | c | c | c | c} 
 & MLE & $\hat {\Sigma}_\J$ & $\hat{\Sigma}_\I$ & $\hat{\Sigma}_{b,p}$ & 
 $\hat{\Sigma}_{b,np}$ & $\hat{\Sigma}_{nb,p}$ & $\hat{\Sigma}_{nb,np}$   \\
\hline
{\tt NFL} MSE & .0591 & .0525 & .0499 & .0491 & .0585 & .0491 &  .0585 \\
\% better & - & 11.1\% & 15.5\% & 16.8\% & 0.9\%  & 16.8\% & 0.9\%\\
{\tt NBA} MSE & .0104 & .0098 & .0098 & .0095 & .0099 & .0094 & .0099 \\
\% better& - & 5.2\% & 5.2\% & 8.8\% & 4.7\% & 9.1\% & 4.7\% \\
\end{tabular}
\includegraphics[width=\columnwidth]{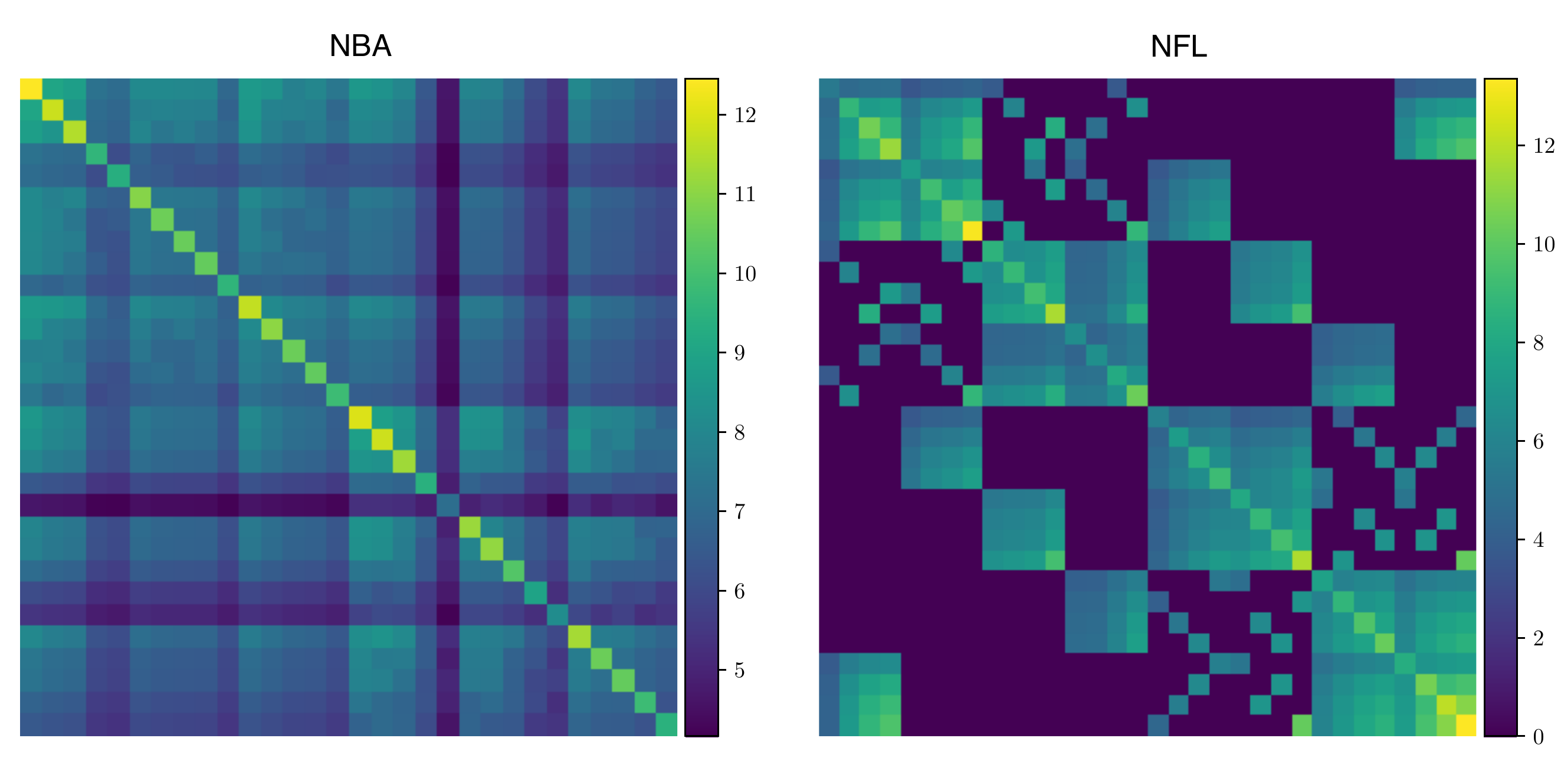} \\ 
\includegraphics[width=.9\columnwidth]{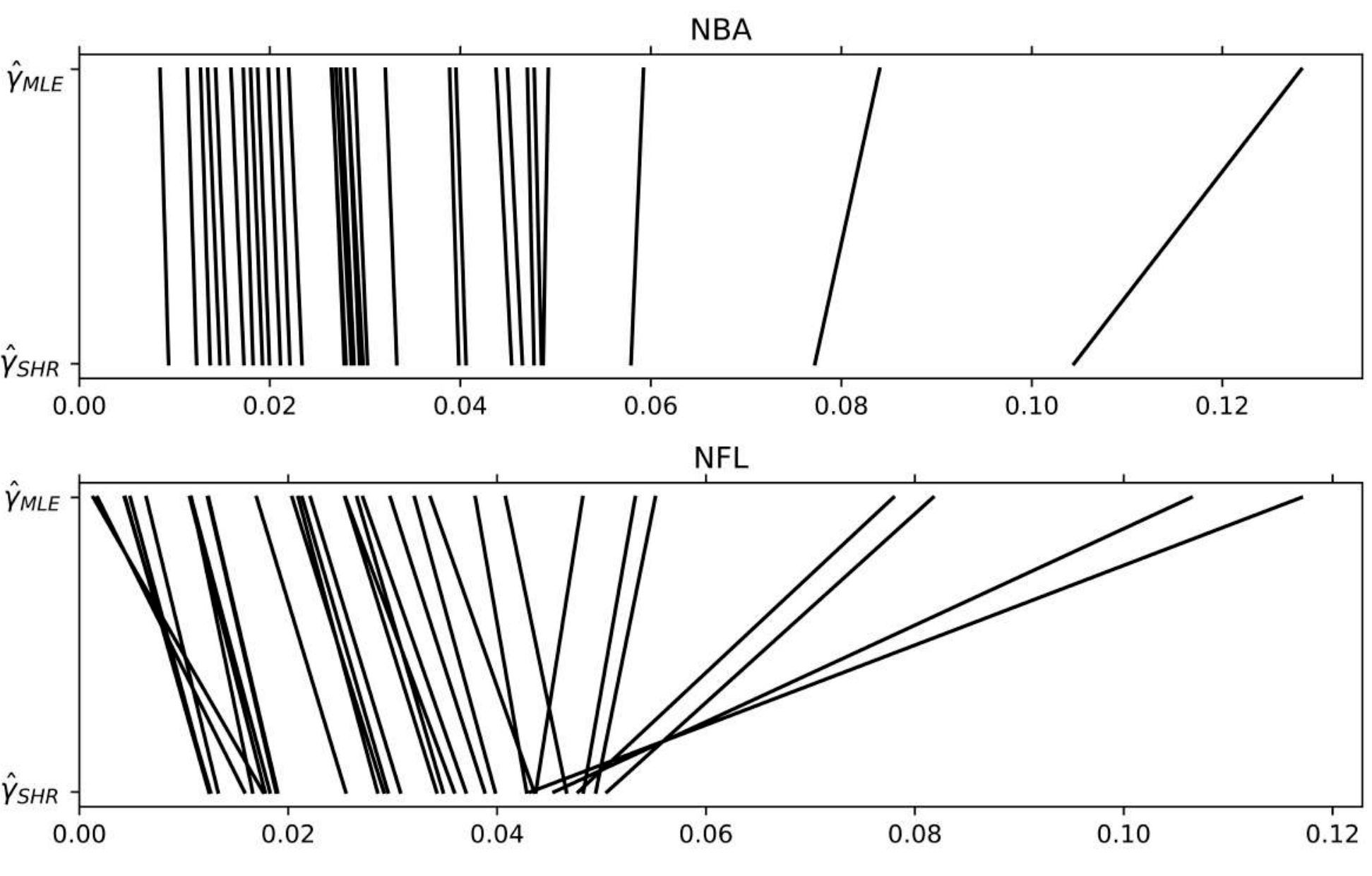} \\
\caption{Results for the \nfl and \nba datasets with conference structure. Top: MSEs and percentage improvement in MSE over the MLE for win percentages in the two datasets with different estimated covariance matrices. For shrinkage using the parametric block bootstrap ($\hat{\Sigma}_{b,p}$), most naturally suitable in this setting, the MSE improves by 16.8\% on the NFL data. The NFL data consists of many fewer games per team and is much less strongly connected than the NBA data. This difference is reflected in the Fisher Information matrix (middle) as well as the pattern of shrinkage (bottom).}
\label{sportsballbro}
\end{center}
\end{figure}

 Because we cannot measure MSE with respect to unknown ground truth parameters, we instead note that under the MNL model, teams facing the same schedule of opponents have a win percentage that is simply a scaling of their quality parameter. The schedules are not identical for e.g.~the NFL, but we accept predicting win percentage it as a suitable proxy for accuracy.
 
 We use $2$-fold cross-validation to evaluate our shrinkage estimator, averaged across $10$ runs, which evaluates how well we can predict one half of the games from another half. We use only 2 folds both because this gives us smaller training sets, the data realm we which to emphasize, and because it gives us larger test sets.

We fit $\gmle$ for an MNL model and then compute $\hat{\gamma}_{SHR}$ with various choices of the covariance matrix estimator: using observed Fisher Information (we refer to this as $\hat{\Sigma}_{\J}$), expected Fisher Information ($\hat{\Sigma}_{\I}$), parametric and non-parametric blocked bootstrap ($\hat{\Sigma}_{b,p}$ and $\hat{\Sigma}_{b,np}$ respectively), and the parametric and non-parametric (non-blocked) bootstrap ($\hat{\Sigma}_{nb,p}$ and $\hat{\Sigma}_{nb,np}$ respectively). 

In Figure~\ref{sportsballbro} we observe that for the blocked bootstraps $\hat{\gamma}_{SHR}$ does significantly better in out-of-sample win percentage prediction than $\gmle$ on both the \nfl and \nba dataset. We further see that while the Fisher information estimations for $\Sigma$ give strong gains, the best improvements come from bootstrap based estimators. We note that while the decision to use the blocked or non-blocked bootstrap has little impact on the effectiveness of the shrinkage, the parametric bootstrap is significantly more effective than the non-parametric bootstraps, particularly on the \nfl data. This is because the non-parametric bootstrap is unable to find the variance when the data is as sparse as the \nfl dataset, where many pairs of teams have played only one game that is then fixed by all non-parametric bootstraps. 

The difference in improvement between NFL and NBA is due to the NFL teams playing relatively few games and the conference structure being less well connected. This is reflected in the Fisher Information matrix (Figure \ref{sportsballbro} middle) and can be seen starkly when we plot the changes in estimated skill levels for each team between $\gmle$ and $\hat{\gamma}_{SHR}$ (Figure \ref{sportsballbro} bottom). In the NFL we see that the shrinkage changes the ordering of some teams while in the NBA data shrinkage does little beyond shrink the estimate of one or two extreme teams slightly. 

Although we have focused on improvement in win rate prediction (Figure~\ref{sportsballbro}), we still observe improvements in mean squared error on individual matchups (equivalently, applying the Brier scoring rule to probabilities). At the level of matchups we observe MSE improvements for the \nfl dataset of 5.4\% and 7.2\% for the Fisher shrinkage with $\hat \Sigma_\J$ and $\hat \Sigma_\I$ respectively, and MSE improvements with the bootstrapped estimators of 9.2\% for both $\hat \Sigma_{b,p}$ and $\hat \Sigma_{nb,p}$ and of 1\% for $\hat \Sigma_{b,np}$ and $\hat \Sigma_{nb,np}$. These matchup results further highlight, in addition to the win percentage results in Figure~\ref{sportsballbro}, that the parametric bootstrap is important for capturing the variance in pairwise matchups that only occur once in the data, and that the Fisher shrinkage is more effective than the non-parametric bootstrap methods but not as effective as the parametric bootstrap methods. We observed similar patterns on the \nba dataset, but because there are so many more matchups in that dataset (42 games per team in the training set), we see gains of less than 1\% on matchup MSE.

\subsection{Increasing survey power}
We now consider a different application: increasing the power of comparison surveys. Here we use a survey dataset from the AllOurIdeas wikisurvey platform \cite{salganik2015wiki}. This data consists of a survey designed by the Washington Post and run on the platform, asking readers which political figure within a pair had ``the worse year in Washington." There are $N=143,704$ comparisons made between $n=67$ figures. The data is available on the AllOurIdeas website, \href{http://blog.allourideas.org/post/2739358388/download-your-data}{http://www.allourideas.org/}. Note that this particular Washington Post wikisurvey is relatively unique among wikisurveys because respondents were recruited through one of the largest news sites in the United States. Most wikisurveys operate in the small-data regime, hundreds or thousands of responses (rather than hundreds of thousands of responses). All training for this dataset used the expected Fisher information $\hat \Sigma_\I$ for shrinkage.

\begin{figure}[t]
\includegraphics[width=\columnwidth]{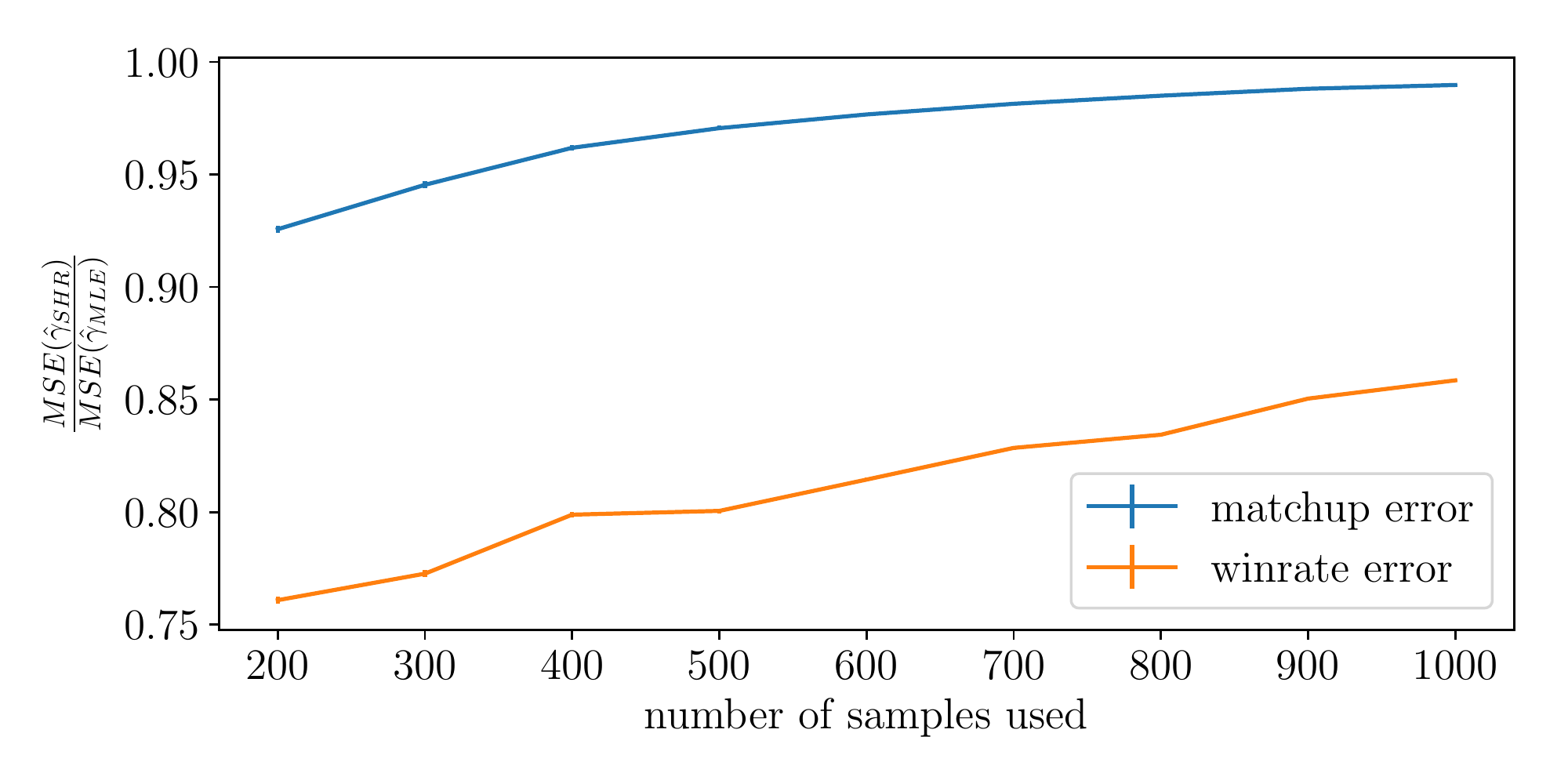}
\caption{The ratio of out-of-sample MSE using $\gshr$ vs. $\gmle$ as a function of the number of training samples for the {\tt AllOurIdeas} dataset.  }
\label{AllOurIdeas-partial}
\end{figure}

We perform our inference on subsamples of the data and evaluate predictive performance on the large held out sample. The results of this examination are shown in Figure~\ref{AllOurIdeas-partial}. Errors are computed using a test set consisting of all data never included in training, and the plot is averaged over 25 shuffles of the data and includes imperceptibly small standard errors. We observe a large reduction in error in terms of predicting pairwise comparisons (matchup error) in smaller samples and a significant reduction in the prediction of probability of an alternative winning a comparison in expectation (win rate) even for relatively large samples. Thus, shrinkage estimators can definitely improve the efficiency of small online surveys.

\subsection{Pitcher/batter matchups}

We now consider a dataset with more comparisons, many more alternatives, and a different structure than either of the two datasets above. The {\tt MLB} data we consider consists of nearly all of the at-bats in the 2016 MLB season. Here we study the use of shrinkage to predict ``on base percentage,'' a motivating example from perhaps the most seminal work on shrinkage \cite{efron1977stein}. 

Because the MLE of an MNL model is defined only on a strongly connected component of a directed comparison graph, we restricted the data only to pitchers and batters in the largest such component. This restriction amounts to removing players who either won or lost all of their matchups, i.e.~batters with no hits, batters with all hits, pitchers giving up no hits, and pitchers giving up all hits. While it seems unfair that a perfect player might be removed, it is often a strong signal that the player is already known by coaching staff to be a weak player, resulting in few matchups. We count walks as hits because they represent a desired outcome for the batter. 

The data was collected from \href{http://www.retrosheet.org/game.htm}{www.retrosheet.org}. Pitchers who also bat are treated as two separate players (their pitching self and their batting self). The unrestricted datasets contains $N$=218,340 at-bats between 1,353 batters and 309 pitchers. The restricted dataset consists of $N$=214,865 at-bats between $n$=1,096 players (787 batters and 309 pitchers). Although restriction drops out about 41\% of the batters, it only throws out 1.6\% of the at-bats. A large number of the removed batters appear to be pitchers serving as batters, which are known to hit rarely and often have only a handful of at-bats across a season. Because the pitcher/batter data reflects a bipartite graph fitting the mold of a Rasch model, we should shrink batters to the mean quality of a batter and pitchers to the mean quality of a pitcher (rather than shrinking both pitchers and batters towards the same mean $u_i = 1/n$). Letting $U_1$ be the set of pitchers and $U_2$ be the set of batters, and let $\gamma^{U_1}$, $\gamma^{U_2}$ be the restrictions of $\gamma$ to $U_1$ and $U_2$ respectively. We thus have 
$u_i = \frac{1}{|U_1|} \sum_{j \in U_1}^n \hat{\gamma}^{U_1}_{MLE,j}$ for $i \in U_1$
and 
$u_i = \frac{1}{|U_2|} \sum_{j \in U_2} \hat{\gamma}^{U_2}_{MLE,j}$ for $i \in U_2$.

In the original study of shrinkage by Efron and Morris, batting averages for batters {\it with exactly 45 at-bats} were estimated using the James--Stein estimator. With the number of at-bats fixed, the shrinkage factor depends only on how much a batter's batting average differs from the population average. The James--Stein estimator can be applied to datasets with multiple sample sizes, in which case the shrinkage factor for a batter depends on both the number of times they've batted as well as how their batting average differs from the population average \cite{said2017}. But such a setting would not account for the differences in strength between the pitchers these batters face, as considered under a Rasch model, and is a method of shrinking averages but not parameters, meaning the output of the James--Stein shrinkage cannot be used directly for matchup prediction. Another key difference between our method and the Efron--Morris method is the consideration of covariance.

\begin{figure}
\centering
\begin{tabular}{ c | c | c | c } 
- & MLE & $\hat \Sigma_\J$ & $\hat \Sigma_\I$   \\
\hline
 MSE & .0209 & .0180 & .0173\\
 \% improvement & - & 13.8\% & 17.2\%
\end{tabular}
\includegraphics[width=\columnwidth]{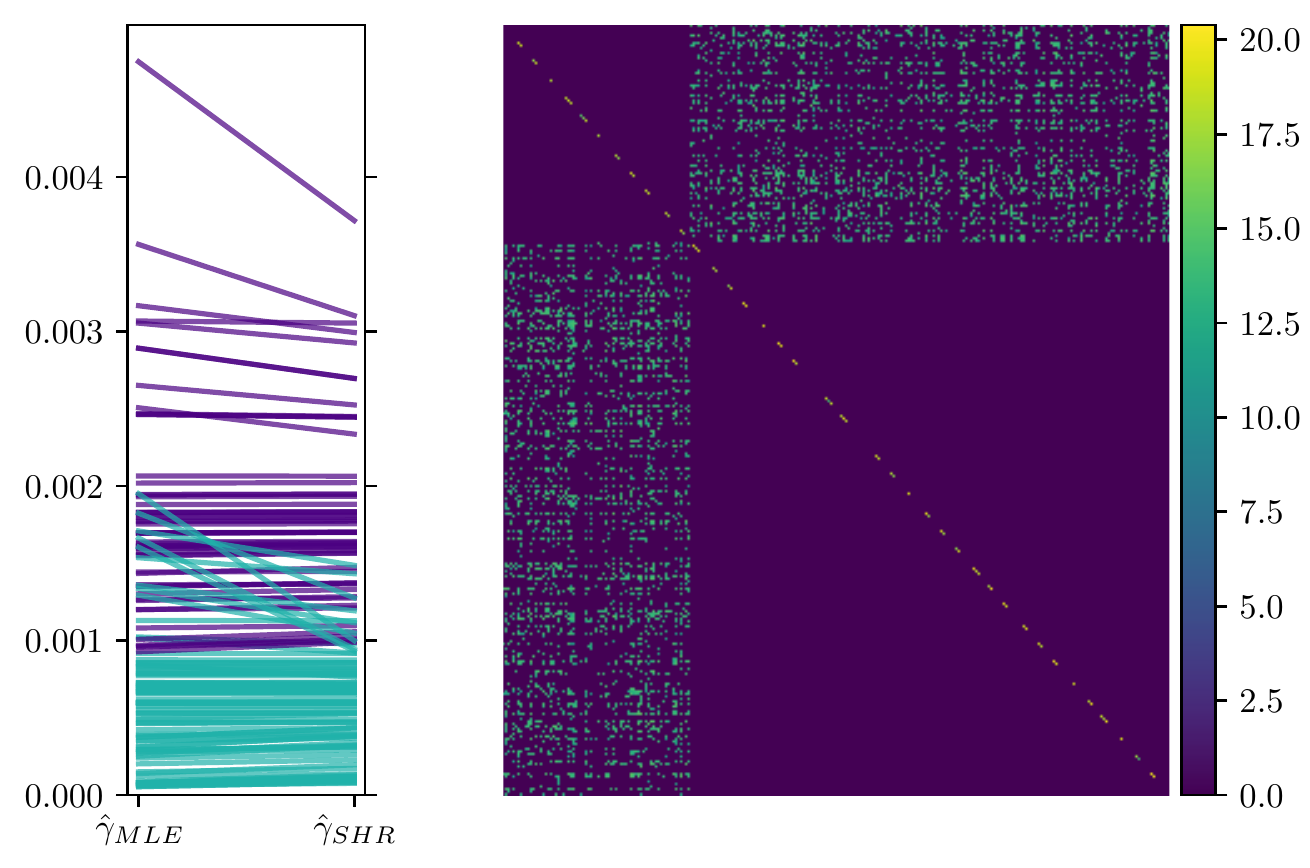} \\
\includegraphics[width=.5\textwidth]{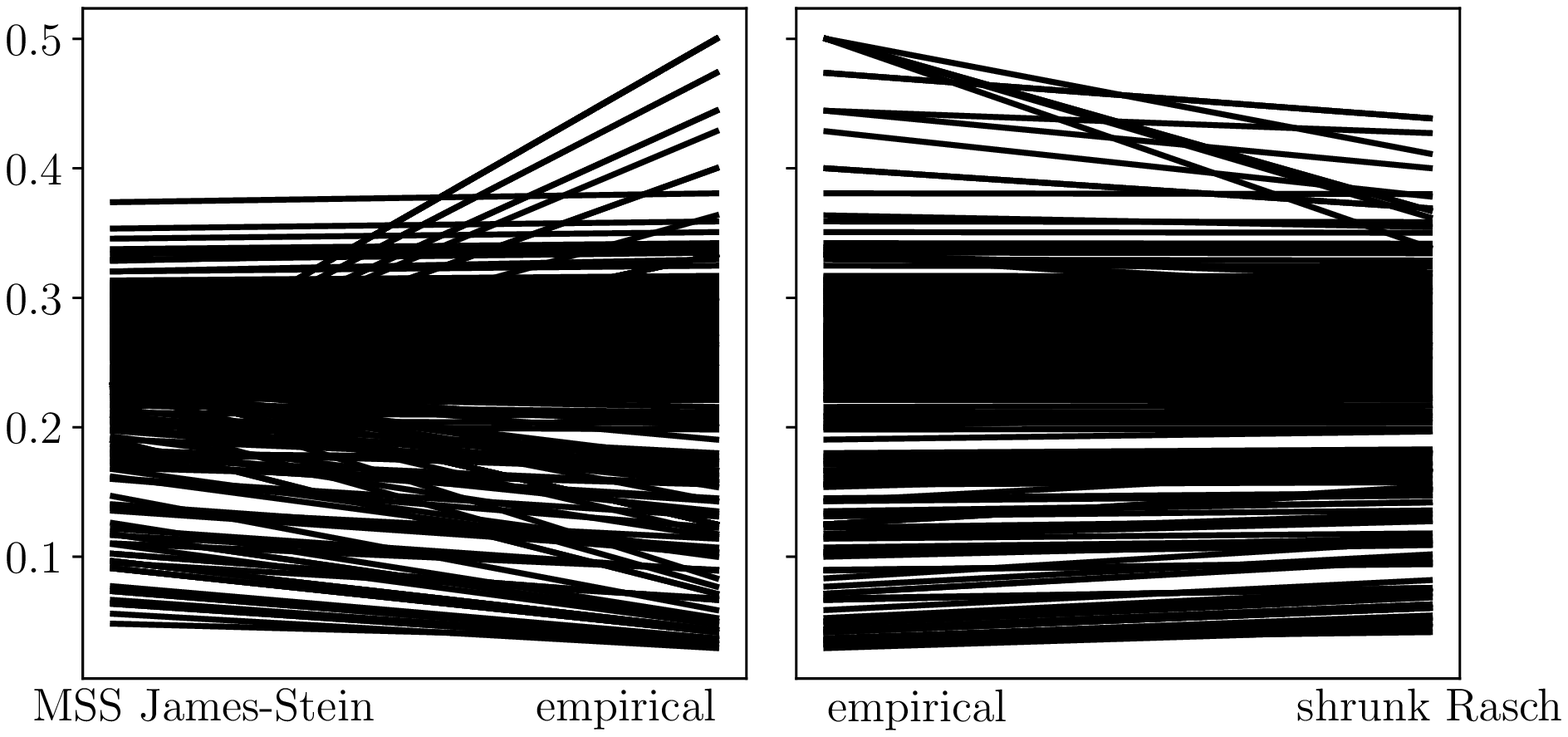}
\caption{Shrinkage for the Rasch model on the {\tt MLB} dataset. The top panel shows the shrinkage (left) of $\hat{\gamma}^{U_1}_{MLE}$ for pitchers (purple) and $\hat{\gamma}^{U_2}_{MLE}$ and batters (teal), and the observed Fisher information matrix ($\hat \Sigma_{\J}$, right) for the matchup data. Bottom: the effect of shrinkage on estimating the average skills of batters. Note that we can see the block structure of pitchers and batters Fisher information and that the Rasch shrinkage contracts $\gamma^{U_1}_{MLE}$ and $\gamma^{U_2}_{MLE}$ to different baselines.} 
\label{fig:batavg}
\end{figure}

In Figure \ref{fig:batavg} we show the empirical batting averages for the 2016 MLB season, the multiple sample size (MSS) James--Stein shrunk estimates, and the batting averages when using the Rasch shrinkage developed in this work. We see that Rasch shrinkage is able to account for the strength of pitchers that the batters are facing -- and shrinkage on those pitchers -- while direct shrinkage of observed batting averages does not. We find through $20$-fold cross validation where we train on 5\% of the data and test on the remaining 95\% that shrunk estimates improve the prediction player win percentages (on-base percentage for batters, one minus this for pitchers) by 13.8\% with the observed Fisher information and 17.2\% with the expected Fisher information in terms mean squared error, relative to MLE estimates. We train on the smaller dataset in order to highlight the data realm where shrinkage is most effective. For this dataset we opt to use Fisher information because bootstraped datasets is almost never strongly connected and resampling all 214,865 at-bats enough times to sufficiently estimate the covariance is intractable. Our inclusion of these estimates highlights the ability of Fisher information-based shrinkage to handle datasets that may be unreasonable or unruly to bootstrap.

\section{Conclusion}

In this work we examined how the covariance structure of parameter uncertainties for choice models can be used to derive shrunk estimates of the parameters. Estimating the covariance structure is itself a subtle task requiring consideration of the joint effects of randomness in comparison outcomes as well as randomness from the distribution of how comparisons are made. We developed four methods for bootstrapping comparison data based on whether we use bootstrapping to resample choices with or without blocking and non-parametrically or parametrically. We also discussed the use of observed and expected Fisher information matrices to estimate covariance, giving faster estimation and avoiding concerns about bootstrapped data being strongly connected, although theoretical guarantees for these estimates are typically asymptotic in the number of data, and shrinkage is most effective when there are fewer samples. 

We showed how to turn both bootstrapped estimates of covariance as well as estimates based on the Fisher information matrix produce shrinkage estimators for the quality parameters of items. We showed through a variety of empirical and synthetic datasets that the estimation provides improved inference, especially on sparse and ill-connected data. We found that when feasible, shrinkage from bootstraps performed the best, though we still saw significant increases in performance using the more quickly estimated Fisher information matrices.

There are several interesting directions for future work based upon our findings here, including deriving the shrinkage for more complex choice models involving higher-dimensional embeddings of alternatives, using more complex models for the distribution of matchups to improve out-of-sample prediction when the comparisons in the choice data are from a different distribution than the test data, and developing a deeper theoretical understanding of the relationship between the distribution of the matchups and expected improvement in MSE provided by shrinkage. 

Our research is an example of a larger trend of constructing specialized regularization procedures for important special cases where standard procedures are inappropriate (e.g. instrumental variable analysis \cite{belloni2012sparse,peysakhovich2017learning}, causal inference \cite{johansson2016learning}, heterogeneous treatment effect estimation \cite{athey2016recursive}). This trend is particularly pronounced in the social and behavioral sciences where analyses typically focus on MLE-based estimators but recent work has begun to show the promise of more modern statistical and machine learning techniques \cite{chernozhukov2016double,epstein2016good,peysakhovich2017using,kleinberg2017theory,peysakhovich2017group,kunzel2017meta}. Given the centrality of choice models in social science we hope our results contribute to this important endeavor.

\appendix
\section{Appendix: Efficiently \\ estimating $\gamma$ with priors}
\label{sec:appendix}
Bootstrapped datasets are not always strongly connected. In this appendix we discuss how to include a Dirichlet prior on the quality parameters $\gamma$ of an MNL model so that the MLE exists. Adding a prior is equivalent to adding a small weighted ``choice" to the data of each alternative from the full set.

The iterative Luce spectral ranking (I-LSR) algorithm introduced in \cite{maystre2015fast} is a both computationally and statistically efficient algorithm that we use to compute the MLE given the MNL model parameters given choice data $\D$. It relies on iteratively estimating $\gmle$ as the stationary distribution of a continuous time Markov chain (CTMC) whose rates are a function of the current estimate and the data $\D$. Because this amounts to solving a linear system and the system is sparse for pairwise data but becomes dense when adding the prior ``data,'' we show here how to efficiently solve for $\gmle(\epsilon)$, the MLE under the prior using a sparse linear system. 

Recall that $M^\D$ is a matrix where $M^D_{ij}$ is the number of times $i$ beats $j$ in $\D$. Let $\ep$ be a length $n$ vector with all entries $\epsilon$. From the Markov chain interpretation we have for $\gamma=\gmle$ that 
\begin{eqnarray*}
\sum_{j \neq i} \left(\frac{M^\D_{ji}}{\gamma_i+\gamma_j}+\epsilon_{j}\right)\gamma_i = \sum_{j \neq i} \left(\frac{C_{ij}}{\gamma_i+\gamma_j}+\epsilon_i\right)\gamma_j,
\end{eqnarray*}
which makes $\gamma$ the solution to the balance equations of the underlying CTMC, $\gamma^T Q = 0$ where 
\begin{eqnarray*}
Q_{ij} = \frac{M^\D_{ji}}{\gamma_i+\gamma_j}+\epsilon_{i},\ 
Q_{ii} = -\sum_{j \neq i} Q_{ij} = -(n-1)\epsilon_i- \sum_{j \neq i} \frac{M^\D_{ij}}{\gamma_i+\gamma_j} .
\end{eqnarray*}

Let $\tilde Q$ be the rate matrix of the CTMC corresponding to the original matrix. Then $Q = \tilde Q - n \diag(\ep) + \ep\1^T$ where $\1$ is a column of ones. So $Q^T \gamma = \tilde Q^T \gamma - n (\ep*\gamma) + ||\gamma||_1 \ep$ where $\ep * \gamma$ is a column vector of element wise multiplication and for $||\gamma||_1 = 1$ (by assumption) we have $\tilde Q^T \gamma = n \ep*(\gamma-1).$

Noting that $\tilde Q$ is a function of $\gamma$, we simply apply the iterative method in the I-LSR algorithm, solving for the iterates of $\gamma$ with the $\tilde Q^T \gamma = n \ep*(\gamma-1)$ rather than $\tilde Q^T \gamma=0$. None of the guarantees of the algorithm change because we have shown that this system is equivalent to the system solved by I-LSR under $Q$. The advantage here is that $Q$ is always dense, regardless of the sparsity of $M$, so when $\tilde Q$ is sparse this equivalent system is still fast to solve. 

\bibliography{pushrink} 
\bibliographystyle{acm}

\end{document}